\documentclass[11pt]{article}

\usepackage[utf8]{inputenc}
\usepackage[T1]{fontenc}
\usepackage[english]{babel}
\usepackage{lmodern}
\usepackage[margin=0.9in]{geometry}
\usepackage{graphicx}
\usepackage{booktabs}
\usepackage{multirow}
\usepackage{tabularx}
\usepackage{array}
\usepackage{amsmath,amssymb}
\usepackage{xurl}
\usepackage[hidelinks,hypertexnames=false]{hyperref}
\usepackage{enumitem}
\usepackage[section]{placeins}
\usepackage{float}
\usepackage{microtype}
\usepackage{xcolor}
\usepackage{caption}

\setlist[itemize]{leftmargin=*,topsep=3pt,itemsep=2pt}
\setlist[enumerate]{leftmargin=*,topsep=3pt,itemsep=2pt}
\hypersetup{
  pdftitle={SAGE: Schema-Guided LLMs for Grant Review},
  pdfauthor={Erik Varapaev, Andrei Chetvergov, Stepan Ukolov, Timofei Sivoraksha, Alexander Evseev, Sergey Bolovtsov},
  pdfkeywords={large language models, grant review, human-in-the-loop, schema-guided reasoning, expert alignment, aspect-based evaluation}
}

\begin{document}
\raggedbottom

\begin{center}
{\LARGE\bfseries SAGE: Schema-Guided LLMs for Grant Review\par}
\vspace{0.9em}

{\normalsize
Erik Varapaev\textsuperscript{1},
Andrei Chetvergov\textsuperscript{2,3},
Stepan Ukolov\textsuperscript{2,3}\\[0.25em]
Timofei Sivoraksha\textsuperscript{2,3},
Alexander Evseev\textsuperscript{2,3},
Sergey Bolovtsov\textsuperscript{2,3}\par}
\vspace{0.75em}

{\small
\textsuperscript{1}ITMO University, St. Petersburg, Russia\\[0.12em]
\textsuperscript{2}Russian Presidential Academy of National Economy and Public Administration, Moscow, Russia\\[0.12em]
\textsuperscript{3}Ivannikov Institute for System Programming of the Russian Academy of Sciences, Moscow, Russia\par}
\vspace{0.55em}

{\footnotesize
\texttt{465350@niuitmo.ru}\\[-0.05em]
\texttt{\{chetvergov-as,ukolov-sd,sivoraksha-ta\}@ranepa.ru}\\[-0.05em]
\texttt{\{aevseev-23-01,bolovtsov-sv\}@ranepa.ru}\par}
\end{center}
\vspace{0.35em}

\begin{abstract}
Grant reviewers must apply detailed criteria to application forms, budgets, and supporting documents while producing an assessment that colleagues can inspect. We present SAGE---Schema-guided Aspect-based Grant Evaluation---a system that translates a grant rubric into structured checks and links its judgements to evidence from the application package. We evaluate SAGE in two stages on 35 nonprofit grant applications. A post-factum comparison with 105 reviews from the original competition shows fair ordinal agreement ($\kappa=0.29$). The foundation then conducted a criterion-level re-review after inspecting SAGE, producing 202 assessments. In this assisted round, SAGE reached $\kappa=0.58$ and outperformed a one-prompt-per-criterion baseline ($\kappa=0.33$ on the common subset), with higher rank correlation and lower error. A claim-level audit further identifies confirmed, disputed, and unaddressed parts of the structured draft. SAGE operationalizes the review methodology by producing a detailed, auditable draft for expert correction.

\medskip\noindent\textbf{Keywords:} large language models, grant review, human-in-the-loop, schema-guided reasoning, expert alignment, aspect-based evaluation
\end{abstract}

\section{Introduction}

Grant-making organizations distribute limited resources among many applicants. Their reviewers rarely work from a single, tidy document. A typical application combines a structured form with a budget, letters, reports, links, and other attachments. Reviewers have to find the relevant evidence, apply a multi-criteria rubric, assign scores, and explain those scores in a form that is useful to both the funder and the applicant. Much of the work is repetitive, but the judgement is not: evidence is unevenly distributed and some criteria depend on how several parts of the proposal fit together.

Large language models (LLMs) can help with reading and drafting, but a free-form ``review this proposal'' prompt is a poor fit for the task. The model may prioritize salient text over the rubric's requirements. A polished explanation can also hide omissions: the reviewer cannot easily tell which requirement was checked or where a factual claim came from. Score agreement does not solve this problem, since a plausible score may still rest on an arithmetic error or on evidence borrowed from the wrong criterion.

For grant review, usefulness depends less on whether the model can imitate the language of an expert than on whether its work can be inspected and corrected. This places the task somewhat apart from benchmark grading and short-answer evaluation. The input is a document package, the criteria belong to a particular funder, and there is no single objective label against which every judgement can be checked.

We call the system SAGE (Schema-guided Aspect-based Grant Evaluation). SAGE decomposes the funder's rubric into criteria, subcriteria, and concrete aspects. It asks the model to complete these checks before a criterion score and comment are assembled. The output is meant to be edited: an expert can inspect the cited evidence, reject a claim, or revise a score before the review is finalized.

The study follows two connected lines of inquiry. The first is methodological: we examine how a funder's review methodology can be represented as an executable schema that produces structured, evidence-linked drafts. The second is empirical: we use two evaluation rounds. The first compares SAGE post-factum with reviews from the original competition. In the second, foundation reviewers revisit the same applications with access to SAGE and enter criterion-level assessments; these assessments are used to compare SAGE with a practical criterion-level prompting baseline and to audit the generated claims.

For this purpose, we developed a grant-review taxonomy containing 8 criteria, 37 subcriteria, and 228 atomic aspects, along with an implementation that turns the taxonomy into an editable review artifact. The released materials include the system and evaluation code, run artifacts, anonymized score pairs, and the aggregate analysis.\footnote{Artifact repository: \url{https://anonymous.4open.science/r/SAGE-DEE5/README.md}}

\begin{figure}[t]
\centering
\includegraphics[width=0.98\textwidth]{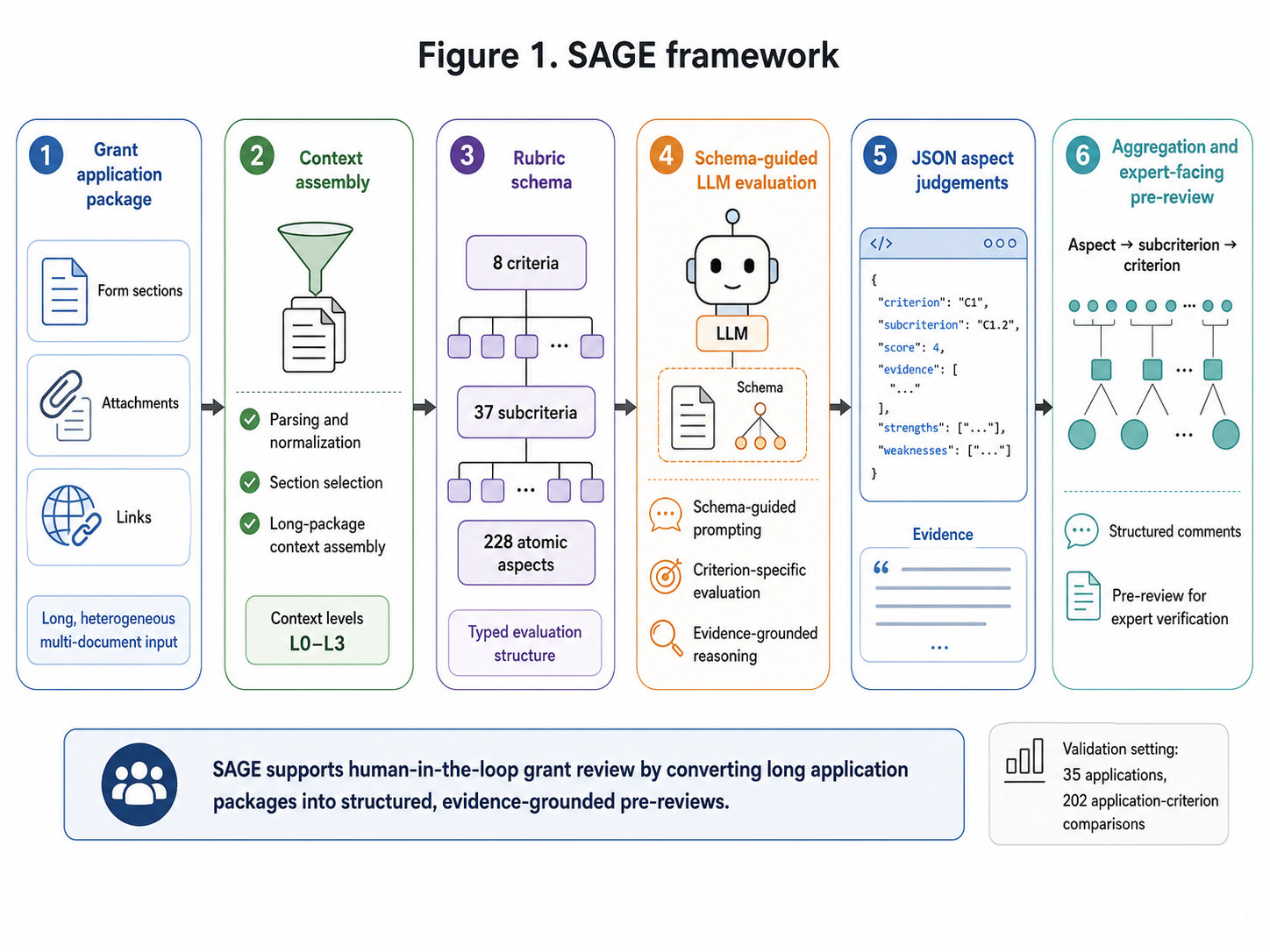}
\caption{SAGE framework. The framework converts a long grant application package into structured, evidence-grounded aspect judgements and aggregates them into an expert-facing pre-review for human verification.}
\label{fig:framework}
\end{figure}

\section{Related Work}

\subsection{Automated Grant Review and Expert Variability}

Grant peer review is known to have limited inter-reviewer reliability. Prior studies report low or moderate agreement between reviewers evaluating the same applications, motivating tools that can make criteria explicit and support reviewer calibration~\cite{pier2018low,mutz2012heterogeneity}. We therefore treat expert ratings as operational reference points and evaluate SAGE as a review-support artifact whose outputs can be inspected, corrected, and compared with expert reasoning.

Most computational work on grant review has focused on analysing reviewer comments or predicting categories in existing reviews. Transformer-based classifiers can identify content categories in grant reports when annotated data are available~\cite{okasa2024supervised}. Such systems, however, typically operate after human reviews have been written and require labelled corpora that many nonprofit grantmakers do not have. SAGE instead supports the review process directly by producing a preliminary structured evaluation from the application package itself.

\subsection{LLM-as-a-Judge and Rubric-Guided Evaluation}

The LLM-as-a-Judge paradigm uses language models to evaluate outputs or documents against criteria~\cite{zheng2023judging,gu2024survey}. Prometheus and Prometheus~2 demonstrate the value of fine-grained, customized score rubrics and user-defined evaluation criteria~\cite{kim2023prometheus,kim2024prometheus2}. Generic judging remains risky for grant review because a global score and fluent explanation can conceal omitted criteria or unsupported evidence.

Providing a scoring guide partially addresses this problem. However, a prose rubric alone does not ensure that all items are checked or every claim is tied to evidence. SAGE operationalizes the rubric as a typed schema. This changes the unit of generation from a free-form review to aspect records that can be parsed, aggregated, and audited.

\subsection{Long Document and Evidence-Grounded Evaluation}

Grant applications are long document packages. Important evidence for one criterion may be distributed across the project description, budget, team profile, outcome table, and attached materials. Retrieval-augmented generation addresses long-document tasks by selecting relevant passages before generation~\cite{lewis2020rag}. SAGE uses a constrained variant: deterministic context assembly guided by the rubric. Each subcriterion is mapped to expected evidence sources, reducing the risk that the model evaluates a criterion using irrelevant context.

The work is also related to aspect-based text analysis, where a document is decomposed into separate dimensions before interpretation~\cite{zhang2022absa}. SAGE extends this idea from sentiment or summarization tasks to multi-criteria expert review: each aspect becomes a typed judgement with a score, evidence, and recommendation.

\begin{table}[ht]
\centering
\caption{Positioning of SAGE against related approaches.}
\label{tab:positioning}
\small
\begin{tabular}{p{27mm}p{43mm}p{49mm}}
\toprule
Approach & Main limitation for grant review & SAGE response \\
\midrule
Rule-based checks & Good for formal validation, but too rigid for substantive expert judgement. & Uses the official rubric as a typed schema while preserving textual reasoning. \\
Classical ML classifiers & Require labelled corpora; their predicted labels do not provide an audit trail. & Works with rubric decomposition and produces evidence-grounded aspect records. \\
Generic LLM judge & Produces fluent global scores, but coverage and grounding are hard to verify. & Forces criterion-by-criterion JSON outputs and downstream claim audit. \\
RAG assistant & Can retrieve evidence, but may not apply all rubric aspects consistently. & Uses deterministic context assembly tied to subcriteria and aspects. \\
\bottomrule
\end{tabular}
\end{table}

\section{Task and Data}

\subsection{Problem Formulation}

Let $D$ be an application package and $R$ be a rubric. The goal is to produce a preliminary review artifact $Y$ that contains scores, evidence-grounded statements, weaknesses, strengths, and recommendations for each criterion. The output should satisfy three requirements. \emph{Coverage}: every required rubric component should be addressed. \emph{Grounding}: generated claims should refer to evidence in the application package or explicitly mark missing evidence. \emph{Expert alignment}: criterion-level scores and claim-level statements should be comparable with human expert judgements.

The target output is a draft review that an expert can audit and amend. Final funding decisions may additionally depend on organizational context, portfolio composition, and information outside the submitted package.

\subsection{Rubric and Validation Data}

The rubric contains eight high-level criteria. Each criterion is decomposed into subcriteria and atomic aspects. An aspect corresponds to a concrete check, such as whether the problem is supported by evidence, whether the target group is specified, whether the budget is consistent with planned activities, or whether expected outcomes are measurable. In total, the schema used in this study contains 8 criteria, 37 subcriteria, and 228 atomic aspects.

The broader empirical archive contains 849 unique grant applications, 149 unique experts, and 2,547 expert--application reviews; it informed rubric analysis and schema design. The validation sample contains 35 applications. The first evaluation round uses three reviews per application from the original competition. These records predate SAGE and yield 615 expert--application--criterion comparisons and 105 weighted application scores. Because the original reviewers differed in their assessments and wrote comments for operational decision making, the foundation later conducted a second, criterion-level review of the same applications. Reviewers inspected the SAGE drafts and entered scores and comments for the applicable criteria, producing 202 numeric assessments. Table~\ref{tab:data} summarizes both rounds.

\begin{table}[ht]
\centering
\caption{Data used in schema design and validation.}
\label{tab:data}
\small
\begin{tabular}{p{47mm}p{17mm}p{54mm}}
\toprule
Level & Count & Role in the study \\
\midrule
Historical applications & 849 & Rubric analysis and schema design \\
Unique experts & 149 & Expert variability context \\
Expert-application reviews & 2,547 & Background expert archive \\
Validation applications & 35 & Shared sample for both evaluation rounds \\
Original reviews per application & 3 & Independent competition records \\
Original expert--application pairs & 105 & Weighted score-level comparison \\
Original criterion comparisons & 615 & Criterion-level historical comparison \\
Assisted criterion assessments & 202 & Foundation re-review after inspecting SAGE \\
Extracted SAGE claims & 1,584 & Audit against assisted expert comments \\
Rubric hierarchy & 8/37/228 & Criteria, subcriteria, and aspects \\
\bottomrule
\end{tabular}
\end{table}

The application materials and original expert comments are confidential. The repository therefore contains anonymized pair-level scores, aggregate tables, prompt and schema skeletons, evaluation scripts, and synthetic examples, but no raw applications or comments.

\subsection{Unit of Analysis}

The two rounds use different units. In the independent round, competition-track weights produce 105 application scores; the criterion table uses all 615 original reports. In the assisted round, the unit is an application--criterion pair, giving 202 expert assessments and 201 pairs with valid scores from both systems. Bootstrap intervals are clustered by application.

SAGE works at a still finer level: each criterion comment combines multiple claims supported by aspect records. The claim audit compares 1,584 extracted SAGE claims with the assisted expert comments. Score metrics describe rating alignment, while the audit shows which parts of the draft the expert comment supports, disputes, or leaves unaddressed.

\section{SAGE Framework}

\subsection{Overview}

Figure~\ref{fig:framework} shows the four parts of SAGE: context assembly, aspect-level generation, aggregation, and expert verification. The rubric is represented as a schema that determines what the model must inspect. This keeps the system close to the funder's methodology and makes omitted checks visible.

\subsection{Aspect Schema}

Let $R=\{c_1,\ldots,c_8\}$ be the set of high-level criteria. Each criterion $c_i$ is decomposed into subcriteria $S_i=\{s_{i1},\ldots,s_{ik}\}$, and each subcriterion contains atomic aspects $A_{ij}=\{a_{ij1},\ldots,a_{ijm}\}$. For each aspect, SAGE specifies relevant application sections, expected evidence, checks to be performed, and a structured output schema.

A simplified aspect-level record contains fields for the criterion, subcriterion, aspect identifier, score, confidence, evidence, strengths, weaknesses, recommendation, and rationale. This structure makes the output machine-readable and enables downstream aggregation and audit. It also encourages explicit missing-evidence markers and limits inference from project intent.

\subsection{Context Assembly}

SAGE constructs the context for each criterion from predefined document layers. It begins with the relevant form sections and then adds attachment text and link summaries when they are available. Restricting the prompt to expected sources keeps it shorter and reduces accidental mixing between criteria.

The mapping also makes the treatment of applications more consistent: the same subcriterion receives the same kinds of evidence each time. If a section or attachment is absent, that absence is passed to the model instead of being filled by inference. Because context selection is recorded separately from generation, a reviewer can distinguish a model error from an extraction or retrieval error.

\subsection{Schema-Guided Generation}

For each subcriterion, SAGE builds a prompt containing the expert role, criterion definition, aspect checklist, relevant context, evidence requirements, scoring rules, and required JSON schema. The model is instructed to distinguish unsupported claims from missing information and to avoid external assumptions. The output is parsed and validated. Invalid JSON is repaired only at the syntactic level; unsupported or empty fields remain visible to the reviewer.

The runtime can be summarized as:
\begin{enumerate}
    \item collect criterion-specific context from the application package;
    \item instantiate the schema-guided prompt for each subcriterion;
    \item call the LLM and parse the JSON output;
    \item validate required fields and retry syntactic failures;
    \item aggregate aspect records into criterion-level scores and comments;
    \item compare criterion-level results with expert judgements.
\end{enumerate}

\subsection{Aggregation}

Aspect-level judgements are aggregated into subcriterion and criterion scores. Let $x_{ij}$ be the normalized score for subcriterion $j$ under criterion $i$ and $w_{ij}$ be its weight. The criterion score is:

\begin{equation}
    score(c_i) = \frac{\sum_j w_{ij}x_{ij}}{\sum_j w_{ij}}.
\end{equation}

The final explanation is assembled from structured aspect records linked to the original document. This makes the final comment traceable to aspect-level evidence. A reviewer can therefore inspect whether a score is driven by a missing document, a weak causal mechanism, an inconsistent budget, or another specific aspect.

\subsection{Reviewer-Facing Artifact}

Reviewers receive both the criterion result and the records behind it. Table~\ref{tab:artifact} shows a simplified synthetic record; an actual criterion comment draws on several records of this kind. An expert can correct the cited evidence, reject a claim, change the score, or remove an irrelevant recommendation without rewriting the whole review.

\begin{table}[t]
\centering
\caption{Simplified synthetic example of a SAGE aspect record.}
\label{tab:artifact}
\small
\begin{tabular}{p{30mm}p{89mm}}
\toprule
Field & Example content \\
\midrule
Criterion & Budget optimality \\
Aspect & Costs are linked to planned activities and target group size \\
Evidence & Budget lines mention staff costs and activity materials; no explicit unit-cost justification is provided. \\
Judgement & Partially supported: the budget is connected to the project plan, but cost assumptions are not fully verifiable. \\
Recommendation & Ask the applicant to add unit-cost explanation and participant-count assumptions. \\
Expert action & Confirm, dispute, edit evidence, or adjust score \\
\bottomrule
\end{tabular}
\end{table}

\section{Evaluation Design}

\begin{figure}[t]
\centering
\includegraphics[width=0.98\textwidth]{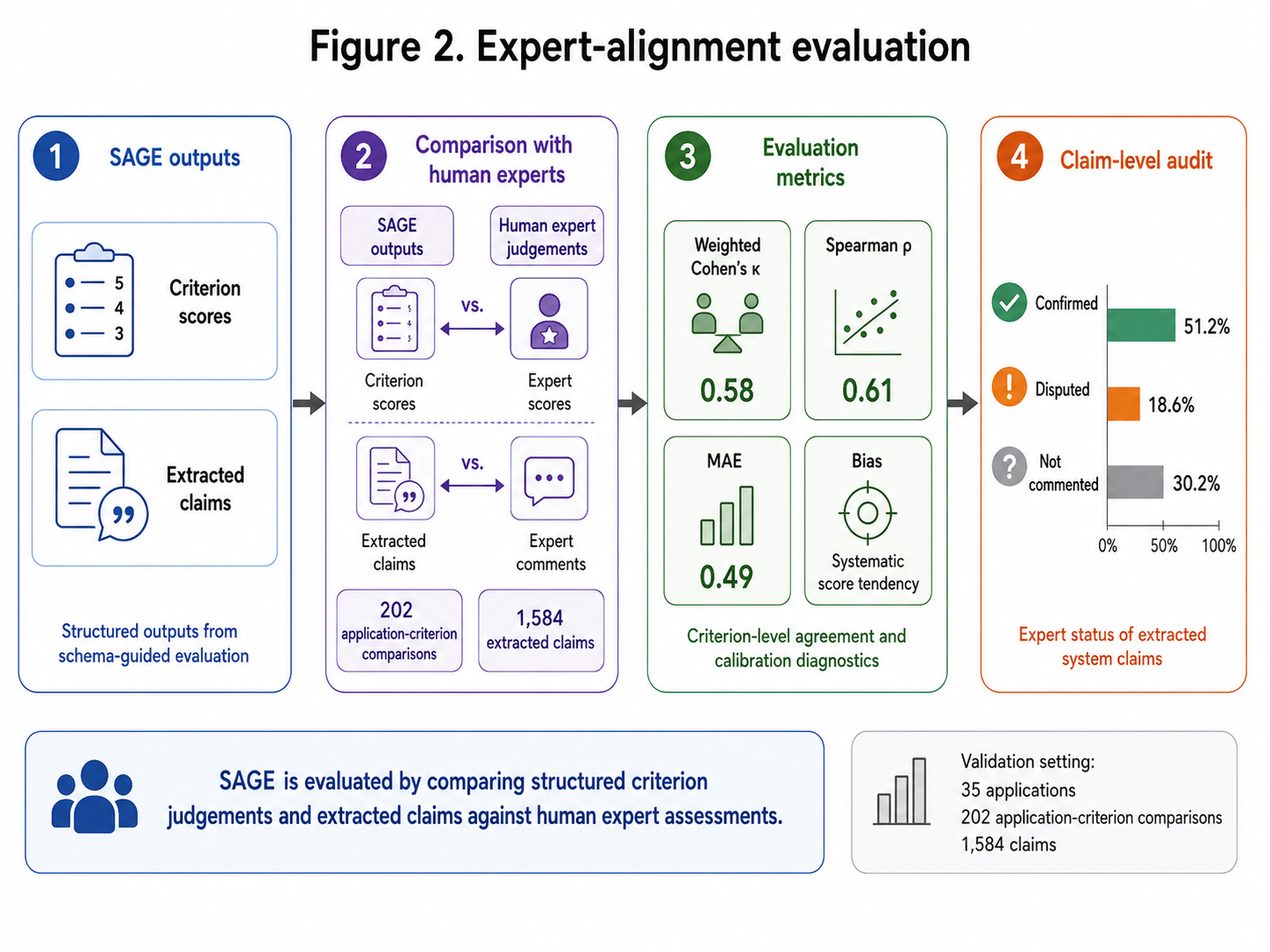}
\caption{Assisted re-review evaluation. Reviewers inspected SAGE and then entered criterion-level scores and comments; extracted SAGE claims were audited against those comments.}
\label{fig:evaluation}
\end{figure}

The evaluation has two rounds. The independent round measures post-factum alignment with decisions made during the original competition. The assisted round captures a later foundation re-review in which reviewers inspected SAGE before entering criterion scores and comments. Figure~\ref{fig:evaluation} summarizes the score and claim analyses for this second round.

\subsection{Two-Round Expert Protocol}

In Round~1, the reference records come from the original grant competition and predate the SAGE run. Each application has three expert reviews. Expert criterion scores were stored as proportions from 0.2 to 1.0 and mapped to the corresponding 1--5 categories. SAGE criterion scores are rounded before the track-specific competition weights are applied. This produces 105 weighted expert--application pairs. Depending on the track, each review contains five or six applicable criteria, giving 615 criterion comparisons.

In Round~2, foundation reviewers revisited the 35 applications with the SAGE criterion drafts visible. They recorded a score and free-text assessment for each applicable criterion, yielding 202 application--criterion pairs. We evaluate SAGE directly against this re-review and run the criterion-rubric baseline on the same applications. One baseline call failed, leaving 201 common pairs for the paired system comparison.

\subsection{Score-Level Metrics}

We compare SAGE scores with expert scores using Spearman rank correlation $\rho$~\cite{spearman1904proof}, quadratic weighted Cohen's kappa $\kappa$~\cite{cohen1968weighted}, mean absolute error (MAE), and bias:

\begin{equation}
    bias = \frac{1}{n}\sum_{i=1}^{n}(\hat{y}_i - y_i), \qquad
    MAE = \frac{1}{n}\sum_{i=1}^{n}|\hat{y}_i - y_i|,
\end{equation}
where $\hat{y}_i$ is a system score and $y_i$ is the corresponding expert score. Quadratic weighting penalizes larger ordinal disagreements more strongly and yields agreement beyond chance on the 1--5 scale~\cite{landis1977measurement}. Scores are rounded only for $\kappa$; Spearman $\rho$, MAE, and bias use the continuous values. We report percentile 95\% confidence intervals from 10,000 application-clustered bootstrap samples (seed 20260713); the same sampled applications are used for paired SAGE--baseline differences.

\subsection{Claim-Level Audit}

Two reviews can assign similar scores for different reasons. An auxiliary pipeline therefore extracts claims from the SAGE criterion comments and matches them with the corresponding assisted expert comments. Each SAGE claim is classified as \emph{confirmed}, \emph{disputed}, or \emph{not commented}. The distinction between confirmed and not commented prevents silence from being counted as agreement. Matching is LLM-assisted and is used as a content diagnostic rather than a manually adjudicated factuality benchmark.

\subsection{Run Configuration}

The SAGE validation used vLLM with a Qwen-family instruction model~\cite{qwen2025technical}. Across 35 applications, it issued 2,974 subcriterion-level requests. All produced parseable outputs; one retry was recorded, and mean latency was 34.1 seconds per request. These values describe run behavior only.

\begin{table}[t]
\centering
\caption{Run-level configuration and operational statistics.}
\label{tab:run}
\small
\begin{tabular}{lr}
\toprule
Quantity & Value \\
\midrule
Validation applications & 35 \\
Original expert--application pairs & 105 \\
Assisted criterion assessments & 202 \\
Baseline common pairs & 201 \\
Extracted claims & 1,584 \\
Subcriterion-level requests & 2,974 \\
Parse success rate & 100\% \\
Recorded retries & 1 \\
Mean time per subcriterion request & 34.1 sec \\
Claim matching & LLM-assisted, not manual \\
\bottomrule
\end{tabular}
\end{table}

\section{Results}

\subsection{Round 1: Independent Historical Agreement}

Table~\ref{tab:overall} summarizes agreement on the 105 weighted application scores. SAGE reaches Spearman $\rho=0.41$ and quadratic weighted $\kappa=0.29$. The MAE is 0.76 points on the five-point scale, and the positive bias of $+0.57$ shows that SAGE scores are generally more generous than the expert ratings. The intervals reflect variation across the 35 applications.

\begin{table}[h]
\centering
\caption{Overall score-level alignment; intervals bootstrap applications.}
\label{tab:overall}
\begin{tabular}{lr}
\toprule
Metric & Estimate [95\% CI] \\
\midrule
Quadratic weighted Cohen's $\kappa$ & 0.29 [0.07, 0.49] \\
Spearman $\rho$ & 0.41 [0.19, 0.60] \\
MAE & 0.76 [0.62, 0.91] \\
Bias, SAGE - expert & +0.57 [+0.41, +0.75] \\
Expert--application pairs & 105 \\
\bottomrule
\end{tabular}
\end{table}

\subsection{Round 1: Criterion-Level Results}

Table~\ref{tab:criteria} reports agreement before criteria are combined into track-weighted totals. The four criteria used in every track show the most consistent ordinal agreement, with quadratic $\kappa$ between 0.29 and 0.42. Optional criteria have fewer observations and show a larger difference between rank association and exact scale calibration.

\begin{table}[t]
\centering
\caption{Agreement by criterion; $\kappa$ is quadratic weighted.}
\label{tab:criteria}
\begin{tabular}{lrrr}
\toprule
Criterion & $n$ & $\rho$ & Quadratic $\kappa$ \\
\midrule
Practice justification & 105 & 0.46 & 0.42 \\
Logical consistency & 105 & 0.46 & 0.38 \\
Result orientation & 105 & 0.28 & 0.34 \\
Budget optimality & 105 & 0.32 & 0.29 \\
Organizational capacity & 90 & 0.33 & 0.02 \\
Evidence quality & 30 & 0.58 & 0.07 \\
Practical experience & 60 & 0.40 & 0.03 \\
Perspective & 15 & -0.37 & -0.06 \\
\midrule
Weighted application total & 105 & 0.41 & 0.29 \\
\bottomrule
\end{tabular}
\end{table}

The criterion results show why the weighted total is the appropriate primary score measure. Some optional criteria preserve rank order while using a different part of the scale, which lowers $\kappa$ despite a positive $\rho$. Perspective is both future-facing and represented by only 15 comparisons. For these criteria, the structured prompts primarily support expert judgement.

\subsection{Round 2: Assisted Re-Review}

The second-round assessments provide a direct view of the intended workflow: reviewers first inspect the structured SAGE draft and then enter their criterion score and comment. Across 202 application--criterion pairs, SAGE reaches Spearman $\rho=0.61$, quadratic weighted $\kappa=0.58$, MAE $=0.49$, and bias $+0.24$. Table~\ref{tab:assisted-criteria} shows that the strongest agreement occurs for result orientation and evidence quality, while the small perspective subset remains difficult.

\begin{table}[h]
\centering
\caption{SAGE alignment in the assisted re-review; $\kappa$ is quadratic weighted.}
\label{tab:assisted-criteria}
\small
\begin{tabular}{lrrrrr}
\toprule
Criterion & $n$ & $\rho$ & $\kappa$ & MAE & Bias \\
\midrule
Evidence quality & 8 & 0.72 & 0.70 & 0.75 & +0.75 \\
Result orientation & 34 & 0.66 & 0.67 & 0.39 & +0.21 \\
Practice justification & 35 & 0.47 & 0.55 & 0.47 & -0.10 \\
Organizational capacity & 30 & 0.58 & 0.55 & 0.41 & +0.34 \\
Budget optimality & 33 & 0.50 & 0.49 & 0.53 & +0.32 \\
Practical experience & 21 & 0.62 & 0.46 & 0.45 & +0.31 \\
Logical consistency & 35 & 0.42 & 0.44 & 0.47 & +0.11 \\
Perspective & 6 & -- & 0.00 & 1.33 & +1.33 \\
\midrule
Overall & 202 & 0.61 & 0.58 & 0.49 & +0.24 \\
\bottomrule
\end{tabular}
\end{table}

\subsection{Round 2: Comparison with the Criterion Baseline}

The practical low-cost baseline extracts predefined form sections and sends one high-level criterion, its full methodology, and the extracted context in a single request. It uses Qwen3.6-35B-A3B, temperature 0.2, guided JSON, and no additional LLM summarization of attachments or links. Unlike SAGE, it does not decompose criteria into 37 subcriteria or 228 aspects. The run produced 279 valid outputs from 280 calls. One missing result leaves 201 application--criterion pairs on which the two systems can be compared against the same assisted assessments. Two zero scores on the baseline's native scale are mapped to one to match the expert 1--5 scale.

\begin{table}[h]
\centering
\caption{SAGE and criterion-rubric baseline in the assisted round.}
\label{tab:baseline}
\small
\begin{tabular}{lrrrrr}
\toprule
System & $n$ & $\rho$ & Quadratic $\kappa$ & MAE & Bias \\
\midrule
SAGE & 201 & 0.61 & 0.58 & 0.49 & +0.24 \\
Criterion rubric baseline & 201 & 0.27 & 0.33 & 0.72 & +0.13 \\
\bottomrule
\end{tabular}
\end{table}

SAGE improves rank association by 0.34 [0.20, 0.49] and quadratic $\kappa$ by 0.25 [0.15, 0.38], while reducing MAE by 0.23 [0.15, 0.31]. All three application-clustered intervals exclude zero. The result supports schema-guided decomposition as the stronger complete configuration for this assisted review setting.

\FloatBarrier
\subsection{Round 2: Claim-Level Audit}

The assisted expert comments also support a finer analysis of the generated review. The matcher extracts 1,584 claims from the SAGE criterion comments. As Table~\ref{tab:claims} shows, 51.2\% are confirmed by the corresponding expert comment, 18.6\% are disputed, and 30.2\% are not addressed. The final category is kept separate because an omitted claim is not evidence of either agreement or disagreement.

\begin{table}[h]
\centering
\caption{Claim-level audit against the assisted expert comments.}
\label{tab:claims}
\begin{tabular}{lrr}
\toprule
Claim status & Count & Share \\
\midrule
Confirmed & 811 & 51.2\% \\
Disputed & 295 & 18.6\% \\
Not commented & 478 & 30.2\% \\
\midrule
Total & 1,584 & 100.0\% \\
\bottomrule
\end{tabular}
\end{table}

\section{Verification Analysis}

We manually inspected comments that the matcher flagged as disputed. Five recurring patterns define useful verification targets (Table~\ref{tab:errors}): evidence crossing criterion boundaries, inferences from missing information, numerical interpretation, generous treatment of partial support, and omitted context. Mapping these patterns to specific aspect records turns a broad request to ``check the model'' into a focused review task.

\FloatBarrier
\begin{table}[t]
\centering
\caption{Review patterns that benefit from targeted expert verification.}
\label{tab:errors}
\scriptsize
\begin{tabular}{>{\raggedright\arraybackslash}p{29mm}>{\raggedright\arraybackslash}p{49mm}>{\raggedright\arraybackslash}p{40mm}}
\toprule
Review pattern & Typical manifestation & Verification value \\
\midrule
Cross-criterion use & Budget or team evidence appears in the rationale for another criterion. & Check the rubric mapping and cited section. \\
Evidence gap & A conclusion refers to a mechanism, document, or partnership absent from the selected context. & Confirm the source or mark the information as missing. \\
Numerical checks & Budget shares, unit costs, or participant counts require calculation across fields. & Recompute values from the structured source fields. \\
Partial support & Project intent receives more weight than the strength of the submitted evidence. & Compare the score with the evidence threshold in the rubric. \\
Context coverage & A relevant attachment is outside the assembled subcriterion context. & Add the source and refresh the affected record. \\
\bottomrule
\end{tabular}
\end{table}

Structured records make these patterns easy to locate. Criterion mixing is visible when a record cites a section associated with another criterion. A strong conclusion paired with an empty or generic evidence field directs attention to its supporting source. Amounts, counts, and percentages can be checked directly against extracted fields. The records therefore provide an efficient verification layer alongside the final comment.

\section{Discussion}

The two rounds answer complementary questions. The original competition records provide an independent external reference: SAGE reaches fair ordinal agreement despite being run only after those reviews were completed. The assisted re-review evaluates the system in its intended position inside the workflow. Here SAGE reaches $\kappa=0.58$ and $\rho=0.61$, with lower error and substantially stronger agreement than the criterion-rubric baseline.

The claim audit explains what the score metrics leave hidden. More than half of the extracted claims are confirmed by the assisted expert comments, while disputed claims identify concrete correction points. The schema therefore contributes both a broad first pass and a traceable interface through which reviewers can add domain and portfolio-level judgement.

The verification patterns also suggest concrete improvements. Deterministic checks can support budget arithmetic and participant counts. Separating missing evidence from negative evidence will make recommendations clearer, while claim-level uncertainty can help reviewers prioritize attention. Each addition builds on the schema without changing the expert-facing workflow.

\section{Deployment and Governance}

In practice, SAGE belongs between document intake and the expert's final assessment. It prepares a consistent first pass and a checklist of evidence, strengths, weaknesses, and open questions. The reviewer verifies the highlighted sources and incorporates contextual knowledge into the final assessment.

The interface should keep evidence fields visible beside each judgement and highlight numerical claims for direct comparison with budgets and participant counts. Claim-level confidence can then serve as a practical cue for review priority, moving low-confidence or high-impact items to the top of the queue.

With verified corrections, disagreements can become calibration material: a funder can inspect ambiguous rubric items and recurrent model overreach. Such use requires access control, audit logs, and separation of model suggestions from final expert-authored decisions.

\section{Reproducibility and Data Availability}

The artifact repository linked in the Introduction supports inspection while protecting confidential application text. It contains the LNCS source, publication figures, prompt and schema skeletons, metric and bootstrap scripts, synthetic examples, anonymized values for both evaluation rounds, and aggregate tables. The released package reproduces every reported score statistic and confidence interval. Raw applications and comments are confidential partner data and cannot be distributed.

\section{Limitations}

The study covers 35 applications from one program and one rubric. The second-round assessments were entered after reviewers inspected SAGE, so the comparison with the post-hoc baseline characterizes this assisted workflow and may include an exposure effect. The two systems also differ in context assembly, model configuration, and computation, making their comparison one of complete configurations rather than an isolated schema ablation. Claim matching is LLM-assisted. A controlled multi-program study with randomized access to SAGE can separately measure its effect on review quality, consistency, and time.

\section{Conclusion}

SAGE augments criterion scores with structured evidence records and explicit correction points. Its scores show fair ordinal agreement with 105 independent reviews from the original competition. In the later assisted re-review, SAGE reaches $\kappa=0.58$ and outperforms the criterion-rubric baseline ($\kappa=0.33$), with higher rank correlation and lower error. The audit of 1,584 claims makes the remaining disagreements local and reviewable.

Together, the two rounds support SAGE as a methodology-aware review assistant: it aligns with prior expert assessments, provides a stronger criterion-level draft than the low-cost baseline in the assisted workflow, and preserves a clear role for expert correction.

\end{document}